\documentclass{article}

\usepackage{PRIMEarxiv}

\usepackage[utf8]{inputenc} 
\usepackage[T1]{fontenc}    
\usepackage{hyperref}       
\usepackage{url}            
\usepackage{booktabs}       
\usepackage{amsfonts}       
\usepackage{nicefrac}       
\usepackage{microtype}      
\usepackage{lipsum}
\usepackage{fancyhdr}       
\usepackage{graphicx}       
\graphicspath{{media/}}     
\usepackage{cite}
\usepackage{amsmath,amssymb,amsfonts}
\usepackage{algorithmic}
\usepackage{graphicx}
\usepackage{textcomp}
\usepackage{makecell} 
\usepackage{booktabs} 
\usepackage{array} 
\usepackage{xcolor}
\usepackage{bm}
\usepackage{multirow}
\usepackage{siunitx} 
\title{FLFL: Federated Latent Factor Learning for Private Recovery of Spatio-Temporal Signals
}
\author{
  Chengjun Yu \\
  Southwest University \\
  Chongqing, China \\
  \texttt{xirimuli@email.swu.edu.cn} \\
   \And
  Di Wu\\
  Southwest University \\
  Chongqing, China \\
  \texttt{wudi1986@swu.edu.cn} \\
     \And
  Yi He\\
  College of William \& Mary \\
  Williamsburg, Virginia, United States \\
  \texttt{yihe@wm.edu} \\
     \And
  Jia Chen\\
  Beihang University \\
  Beijing, China\\
  \texttt{chenjia@buaa.edu.cn} \\
}

\begin{document}
\maketitle
\begin{abstract}

Wireless sensor network (WSNs) stands out as a burgeoning and promising domain in intelligent sensing. Owing to various factors such as sudden sensor malfunctions or deliberate shutdown of partial nodes to save energy, the collected sensing signals from WSNs commonly have massive missing data, leading to adverse effects on subsequent analysis or decision-making. Latent factor learning (LFL) has proven to be highly effective in recovering the missing data for WSNs. However, the existing LFL models require the collected sensing signals to be maintained in one central place like a central server, which is becoming unacceptable for data owners who are getting increasingly privacy-sensitive. To address this issue, this paper innovatively proposes a federated latent factor learning (FLFL) model for privacy-preserving spatio-temporal signal recovery. Its main idea is two-fold: 1) it designs a sensor-level federated learning framework based on LFL, where each sensor only needs to upload gradient information rather than raw data for training a privacy-preserving recovery model, and 2) it incorporates the spatio-temporal correlation into the designed federated learning framework as the regularization constraint to improve its recovery accuracy. With such designs, FLFL can not only accurately recover the missing data of WSNs but also ensure data owners' privacy-preserving of raw data. To evaluate the proposed FLFL model, extensive experiments have been conducted on four real-world WSN datasets. The results demonstrate that FLFL significantly outperforms eight state-of-the-art federated and non-federated signal recovery models in terms of recovery accuracy with privacy-preserving.
\end{abstract}


\section{Introduction}

Wireless Sensor Networks (WSNs) are task-oriented networks composed of multiple sensor nodes that process monitored data through embedded computing modules and transmit it to remote users or sites via wireless networks\cite{yu2022robust}. In the era of big data, WSNs are widely applied in various fields such as smart cities, industrial production, energy, network security\cite{lv2021ai}, bridge engineering\cite{putra2020multiagent}, transportation, and computer vision\cite{wang2020model}, emerging as a frontier technology in intelligent sensing. However, WSNs data are often incomplete due to sensor failures, energy-saving strategies, human errors, and other factors. As WSNs typically collect data at fixed locations and predetermined time intervals over long periods, the incomplete data\cite{b57,b59,b60} exhibit spatial and temporal correlations. These data can be modeled as a low-rank matrix, where rows represent sensor nodes, columns denote time intervals, and elements correspond to recorded WSNs data\cite{48}. Consequently, low-rank matrix approximation (LRMA)\cite{wu2024outlier,47,51,b61,b62} has been widely adopted to recover missing data in WSNs.

Privacy concerns are growing, and regulations are becoming stricter. For example, the GDPR explicitly prohibits collecting, processing, or sharing user data without consent \cite{li2021survey,zhang2021survey}. As a result, data owners are unwilling to exchange raw observations \cite{yang2019federated}. Resource management also discourages data sharing. These factors make traditional centralized modeling—such as Latent Factor Analysis (LFA) \cite{17,22,23,24,43,b5,b7,b21}—unsuitable for wireless sensor networks (WSNs). This is especially true in city‑scale or cross‑institution deployments, where data from different parties are hard to centralize. The core challenge then becomes: under privacy and compliance constraints, and without exposing raw data, how can we model the intrinsic spatio‑temporal correlations of WSNs to achieve accurate missing‑data recovery and efficient representation learning?

Federated learning (FL) has gained significant attention as a privacy-preserving paradigm for collaborative modeling \cite{w20,w21,w22}. FL trains models locally on distributed data and aggregates model parameters or gradients instead of raw samples, thereby extracting shared knowledge \cite{w23,w24}. It has shown promise in recommendation systems\cite{w25,w26}, intelligent medical diagnosis\cite{w27}, and wireless communications\cite{w28}. However, generic FL frameworks fall short for WSNs specific data recovery. In particular, they provide limited support for strong spatio-temporal correlation modeling, and simple federated strategies struggle to exploit node-level spatio-temporal priors. Therefore, WSNs-oriented federated optimization methods are needed that explicitly encode and leverage spatio-temporal priors. Such methods can enable more robust missing-data recovery and more efficient representation learning under privacy-aware settings.

To address this challenge, this paper innovatively propos-es a federated latent factor learning for privacy-preserving spa-tio-temporal signal recovery model, named FLFL, to recover missing data based on partially observed data mixed with privacy-preserving in WSNs. Its main idea is two-fold: 1) We design a sensor-level federated learning framework based on LFL, where each sensor only needs to upload gradient information rather than raw data for training a privacy-preserving recovery model, and 2) FLFL incorporates the spatio-temporal correlation into the designed federated learning framework as the regularization constraint to improve its recovery accuracy. With such designs, it possesses the merits of both high accuracy and privacy-preserving during data recovery. The main contributions of this work are as follows.
\begin{enumerate}
    \item An FLFL model is proposed. It can accurately recover missing data based on partially observed data mixed with privacy-preserving in WSNs.
    \item Theoretical analyses and algorithm design are provided for the proposed FLFL model.
    \item Proofs that gradient information will not be leaked are provided for the proposed FLFL model.
\end{enumerate}

Besides, extensive experiments on four real-world WSNs datasets are performed to evaluate the proposed FLFL model, including comparison with state-of-the-art models and anal-yses of its characteristics. The results verify that FLFL has significantly higher recovery accuracy than the comparison models.

\section{Related Work}
\subsection{Latent Feature Analysis}
The LRMA is widely used for data modeling in WSNs. Because sensors are fixed in space and sample over long horizons at predefined intervals, measurements exhibit inherent temporal and spatial correlations~\cite{29,brtvgs,lrds,30}. Adjacent timestamps and nearby sensors often show similar patterns. LRMA therefore represents WSNs data as a matrix: rows correspond to sensor nodes, columns to time steps, and entries to observed readings. To improve recovery accuracy, training typically embeds spatio\mbox{-}temporal priors~\cite{31,32}. Common choices include neighborhood regularization~\cite{29} and graph-based penalties that exploit the graph structure of many spatio\mbox{-}temporal signals~\cite{33,34,trss}. For example, Qiu et~al.\cite{brtvgs} employs a graph Laplacian to promote smoothness on the network, while Mao et~al.\cite{lrds} introduces a hybrid graph regularizer to capture both global and local smoothness.

However, all the above LRMA-based models are centralized. They cannot preserve the privacy and security of data, which is unacceptable for more and more privacy-sensitive sensors. In comparison, the proposed FLFL model is an FL-oriented LRMA approach, which can not only well incorporate the spa-tio-temporal correlation into LRMA to enhance the accuracy of data recovery, but also well preserve the privacy and security of data in representing WSNs.
\subsection{Federated Learning}
FL is a distributed training paradigm introduced by Google~\cite{18}. It enables multiple participants to jointly train a shared model while keeping raw data local, thereby strengthening privacy protection~\cite{35,36}. In FL, clients update model parameters on-device and share only model updates with a coordinating server, avoiding centralization of personal training data. Privacy concerns have spurred a surge of FL methods. Among them, matrix factorization (MF) within federated frameworks has received particular attention. Chai et~al.\ proposed a secure MF approach under FL, \emph{FedMF}~\cite{FedMF}. Lin et~al.\ developed a federated recommender based on MF with a hybrid imputation strategy, \emph{FedRec}~\cite{Fedrec}. Liang et~al.\ further enhanced \emph{FedRec} by introducing denoising clients to mitigate noise in a privacy-aware manner, \emph{FedRec++}~\cite{fedrec++}. Lin et~al.\ also integrated meta-learning into MF for federated recommendation, \emph{MetaMF}~\cite{MetaMf}. Beyond MF, work has explored broader techniques to improve FL utility under privacy constraints; for example, Wu et~al.\ presented a privacy-preserving federated framework for graph neural network–based recommendation, \emph{FedGNN}~\cite{Fedgnn}.

Nevertheless, when applying the aforementioned federated learning model to WSNs, we observed a lack of sufficient consideration for the inherent characteristics of WSNs data, specifically its spatio-temporal correlation. This resulted in suboptimal performance in data recovery. To address this challenge, we propose a novel model called FLFL. In comparison to conventional models, the FLFL model adeptly incorporates spatio-temporal correlation into the model while preserving data privacy. This clever integration significantly enhances the model's accuracy in data recovery.
\section{Preliminaries}

\subsection{Problem of Data Recovery in WSNs}
Generally, data collected from WSNs is often incomplete due to various factors such as sudden sensor failures, deliberate node shutdowns to conserve energy, human errors, and uncontrollable environmental conditions. The primary challenge in WSNs data recovery lies in reconstructing the missing data using the available observed data. Below, we provide a formal problem formulation.

\textbf{\emph{Definition }1 \emph{(Problem of Data Recovery in WSNs).}} Given a matrix $Y$ from WSNs consisting of $M$ rows and $N$ columns, intended for recording the data generated by $M$ sensors during $N$ time slots. Each element $y_{i,j}$ of $Y$ contains a value corresponding to the data recorded by the $i$-th sensor at the $j$-th time slot. Let $Y_K$ and $Y_U$ denote the known and unknown entry set of $Y$, respectively. The problem of data recovery in WSNs lies in estimating $Y_U$ solely based on the observations $Y_K$\cite{b58}. The value of $Y_K/Y$ is denoted as the sampling rate.
\subsection{LFA-Based Data Recovery in WSNS }
\begin{figure}
    \centering
    \includegraphics[width=0.5\linewidth]{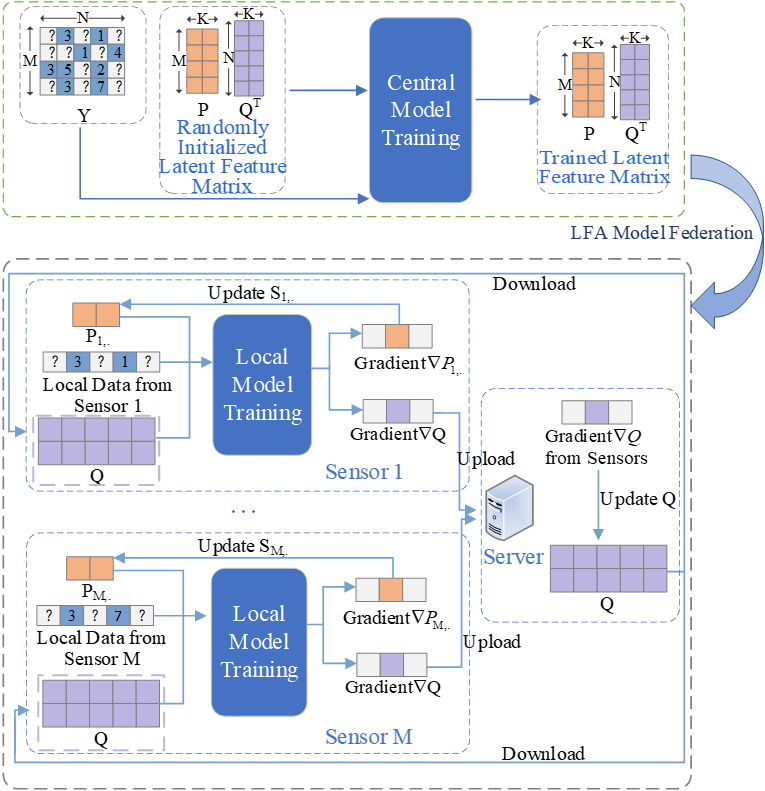}
    \caption{LFA Model Federation in WSNs}
    \label{fig:Federated_LFA}
\end{figure}
\textbf{\emph{Definition }2 \emph{(LFA-Based Data Recovery in WSNs).}} To recover \( Y_U \), the LFA model aims to learn two latent feature matrices \( P \in \mathbb{R}^{M \times k} \) and \( Q \in \mathbb{R}^{N \times k} \), where \( k \ll \min\{M, N\} \)\cite{b66,b67,b68,b69,b70,b73}. The goal is to obtain a rank-\( k \) approximation of \( Y \) by minimizing the discrepancy between the observed entries in \( Y_K \) and their estimates\cite{yuan2020multilayered,luo2020adaptive,b3,b23}. The matrix \( Y \) is approximated as:
\begin{equation}\label{1}
Y \approx PQ^T
\end{equation}
This formulation enables the model to capture the underlying structure of \( Y \) using low-dimensional representations of its rows and columns.

The \( L_2 \) norm is widely used in existing LFA models\cite{20,25,b31,b35,b72}. To avoid overfitting on \( Y_K \), Tikhonov regularization can be integrated into the objective function \( \varepsilon(P, Q) \). The objective function can be formulated as:
\begin{equation}\label{2}
\varepsilon \left( {P,Q} \right) = \left\| {J \circ \left( {Y - P{Q^{\rm T}}} \right)} \right\|_{{L_2}}^2 + \lambda \left( {\left\| P \right\|_F^2 + \left\| Q \right\|_F^2} \right)
\end{equation}
where $\circ$ denotes the Hadamard product, \(J\) is the sampling matrices, \( \| \cdot \|_{L_2} \)  denotes the $L_2$norm of a matrix\cite{21,56,58,b55,b56}, and $||\cdot||_F$ denotes the Frobenius norm of a matrix\cite{59,60,b28}.

\subsection{Federated Learning}
\textbf{\emph{Definition} 3 \emph{(Federated Learning).}} Consider an FL system comprising \( M \) sensors, where each sensor \( i \) possesses a local training dataset \( S_i \) \((i = 1, 2, \dots, M)\). Let \( S \) denote the joint training dataset, representing the union of all local datasets. The goal of FL is to collaboratively train a shared machine learning model based on \( S \) without directly sharing raw data.  

FL can be classified into three categories according to data distribution across the feature and sample spaces: horizontal FL, vertical FL, and federated transfer learning. Given that each sensor in WSNs described in this paper shares data with the same feature space, the FL in this paper belongs to vertical FL, as depicted in Fig.\ref{fig:Federated_LFA}. The training process of the FL system usually contains the following four steps as follows.
\begin{itemize}
 \item Step 1 : The central server initializes the global model and distributes it to all sensors.  
 \item Step 2 : Each sensor trains a local model using its private data and the global model received. After training is complete, instead of sharing raw data, only the computed gradients are sent back to the server.  
 \item Step 3 : The server aggregates the gradients from all sensors to update the global model and redistributes the updated model to each sensor.  
 \item Step 4 : Steps 2 and 3 are iteratively repeated until the model converges. 
\end{itemize}

\section{THE PROPOSED FLFL MODEL}

\subsection{The Framework of FLFL-SSR}\label{}
FLFL model is a collaborative training-shared LFA model in WSNs. Unlike models based on centralized learning, the FLFL model does not require sensors to exchange raw data, thus preserving the privacy of sensor data. Additionally, considering the spatiotemporal correlation inherent in sensor data, the FLFL model incorporates this relationship into its training to enhance its ability to recover data.

\begin{figure}[htbp]
\centerline{\includegraphics[width=\linewidth]{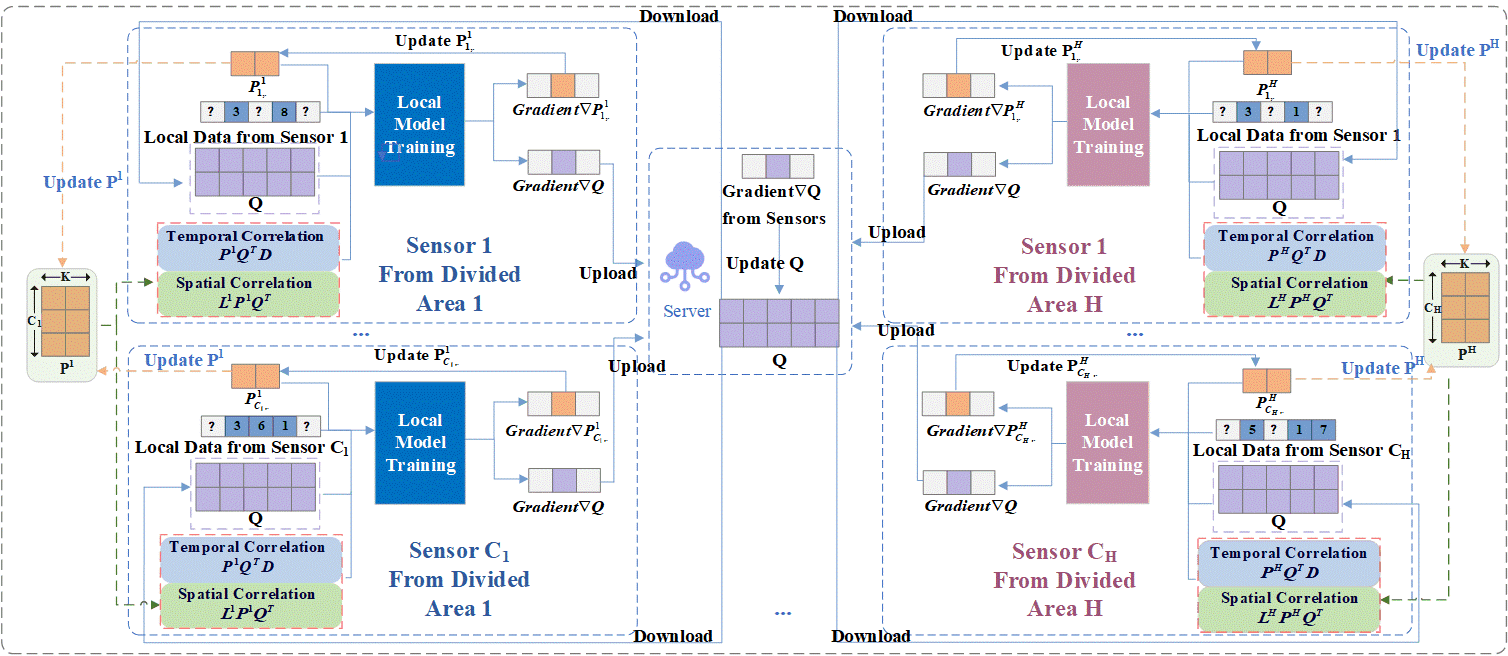}}
\caption{The overall framework of the proposed FLFL model}
\label{fig:FLFL}
\end{figure}

Fig.\ref{fig:FLFL} illustrates the overall structure of the proposed FLFL model. For each sensor $i$, latent feature vectors $\mathbf{P}_{i,:}$ and the necessary raw data during the training process are stored locally. It is noteworthy that sensors located in the same region can share their latent feature vectors. Correspondingly, the server only stores the latent feature matrix $\mathbf{Q}$. In the interaction with each sensor, the system ensures coordination and efficiency through the following five steps:

\begin{itemize}
\item At the initialization stage, the server initializes the latent feature matrix randomly, while each sensor randomly initializes its latent feature vector.
\item Each sensor downloads the latent feature matrix from the server.  
\item Each sensor initiates local training using its local data, the downloaded latent feature matrix from the server, its local latent feature vector, and the local latent feature vectors of other sensors within the same region. 
\item Each sensor updates its own latent feature vector and uploads the gradients to the server.
\item The server updates the latent feature matrix based on the gradients received from each sensor.
\end{itemize}
\subsection{Spatio-Temporal Correlation in FLFL}
The spatio-temporal correlation of WSNs exhibit two fundamental characteristics: spatial smoothness and tem-poral smoothness. Spatial smoothness implies that data col-lected from sensors in close geographical proximity tend to be similar to each other at a given time. Temporal smoothness, on the other hand, signifies that data gathered from the same sen-sors exhibit smooth variations over time.
\subsubsection{Spatial smoothness}
To address the characteristics of sensor data and ensure privacy protection, we partition the space into \( H \) regions. To capture spatial smoothness, we construct an undirected weighted sensor connectivity graph for each region, denoted as \( G_h = (V_h, E_h, W_h) \), where \( h = 1, 2, \dots, H \)\cite{48}. In this graph, \( V_h \) represents the set of vertices corresponding to the sensors within the \( h \)-th region. The number of sensors in this region is \( C_h \), meaning \( |V_h| = C_h \). The total number of sensors across all regions is \( M \), satisfying \( M = \sum_{h=1}^{H} C_h \). Additionally, \( E_h \) denotes the set of edges in the graph \( G_h \), where each edge represents the relationship between a pair of sensors.  

The weighted adjacency matrix \( W_h \) quantifies the connections between sensors, with each entry \( w_{i,i'} \) indicating the strength of the relationship between the \( i \)-th and \( i' \)-th sensors. Based on \( V_h \), we compute the distance matrix \( F_h \) using the sensors’ coordinates, where each element \( f_{i,i'}^h \) captures the Euclidean distance between the \( i \)-th and \( i' \)-th vertices. Leveraging \( F_h \), \( W_h \) is constructed by selecting the top-\( k \) nearest neighbors for each sensor, as follows:
\begin{equation}\label{3}
w_{_{i,i'}}^h = \left\{ \begin{array}{l}
\frac{1}{{{{\left( {f_{i,i'}^h} \right)}^2}}},{\rm{ }}\ if{ \rm{ }}\;  i'\in To{p_k}(i)\\
0,{\rm{     }}\qquad otherwise
\end{array} \right.,
\end{equation}
where \( \text{Top}_k(i) \) represents the set of the top-\( k \) nearest vertices to the \( i \)-th vertex. Furthermore, let \( L_h \) denote the Laplacian matrix of the graph \( G_h \). The matrix \( L_h \) is defined as follows:

\begin{equation}\label{4}
L^h = 
\scalebox{1}{$\displaystyle 
\begin{bmatrix}
sum(w_{1,\cdot}^h) & -w_{1,2}^h        & \cdots & -w_{1,C_h}^h \\
-w_{2,1}^h                  & sum(w_{2,\cdot}^h) & \cdots & -w_{2,C_h}^h \\
\vdots                      & \vdots            & \ddots & \vdots \\
-w_{C_h,1}^h               & -w_{C_h,2}^h      & \cdots & sum(w_{C_h,\cdot}^h)
\end{bmatrix}_{C_h \times C_h}
$} 
\end{equation}
where \( sum(w^h_{i,.}) \), where \( i \in \{1, 2, \dots, C_h\} \), represents the sum of the \( i \)-th row of the weight matrix \( W_h \). To ensure the spatial smoothness of the reconstructed matrix \( G_h \), a regularization constraint is introduced by minimizing \( \|L_h P Q^T\|_F^2 \).

\subsubsection{Temporal smoothness}
Since each row of the WSNs data matrix $Y$ corresponds to the sequential readings of a sensor at fixed intervals, 
the entries along a row generally follow a smooth temporal pattern. 
That is, consecutive elements in the same row tend to exhibit only minor variations. 
Consequently, the differences between adjacent time slots should remain small, and this constraint should also hold for the reconstructed matrix. 
To enforce such smoothness, we introduce the temporal difference matrix $D$ defined as follows:
\begin{equation}
\label{5}
    D = {\left[ {\begin{array}{*{20}{c}}
{ - 1}&0&0&{...}&0\\
1&{ - 1}&0&{...}&0\\
0&1&{ - 1}&{...}&0\\
{...}&{...}&{...}&{...}&{...}\\
0&0&{...}&1&{ - 1}\\
0&0&{...}&0&1
\end{array}} \right]_{N \times N - 1}},
\end{equation}
$\big\| P Q^{T}D \big\|_{F}^{2}$ captures the temporal variation of the reconstructed data. 
Specifically, it encodes the successive differences between adjacent time slots, expressed as $[(\hat{y}_{i,1}-\hat{y}_{i,0})^2,\; (\hat{y}_{i,2}-\hat{y}_{i,1})^2,\; \ldots,\; (\hat{y}_{i,N}-\hat{y}_{i,N-1})^2]$, 
which quantify the changes of sensor $i$ across consecutive intervals.

\subsection{Federated Optimization}
Owing to the decentralized nature of federated learning, each sensor leverages its local data, the latent feature vectors of neighboring sensors within the same region, and the latent feature matrix provided by the server during training. Consequently, the objective function for each sensor \( i \) in the \( h \)-th region can be formulated as follows:
\begin{equation}\label{6}
\scalebox{1}{$\displaystyle 
\begin{split}
\varepsilon \big( p_{i,\cdot}^h,Q \big) = 
& \sum_{\substack{y_{i,j} \in Y_K^i}} \Big( y_{i,j} - p_{i,\cdot}^h q_{j,\cdot} \Big)^2 + \lambda \sum_{\substack{y_{i,j} \in Y_K^i}} \Big( \big( p_{i,\cdot}^h \big)^2 + \big( q_{j,\cdot} \big)^2 \Big) \\
& + z_1 \sum_{\substack{y_{i,j} \in Y_K^i}} \Big( \big( L^h P^h Q^{\mathsf{T}} \big)_{(i,j)} \Big)^2 + z_2 \sum_{y_{i,j} \in Y_K^i} 
    \Big( \big( P^h Q^{\rm T} D \big)_{(i,j)} \Big)^2.
\end{split}
$} 
\end{equation}
where \( Y_i^K \) represents the set of known entries in \( Y \) for sensor \( i \), and \( P_h \) comprises all latent feature vectors of sensors within region \( h \). It is important to note that the Laplacian matrix \( L_h \) varies across different regions. Consequently, the instantaneous loss \( \varepsilon_{h}^{i,j} \) of \( \varepsilon(P, Q) \) for the \( i \)-th sensor at the \( j \)-th time slot in region \( h \), calculated on a single entry \( y_{i,j} \in Y_i^K \), can be expressed as:
\begin{equation}\label{7}
\scalebox{1}{$\displaystyle 
\begin{split}
\varepsilon_{_{i,j}}^{h}={\left( {{y_{i,j}} - p_{_{i,.}}^h{q_{j,.}}} \right)^2}+\lambda\left({{{\left( {p_{_{i,.}}^h} \right)}^2} +{{\left( {{q_{j,.}}} \right)}^2}}\right)
\\+{z}{\left({{{\left({{L^h}{P^h}{Q^{T}}}\right)}_{\left({i,j}\right)}}}\right)^2} + z_2 \Big( \big( P^h Q^{\rm T} D \big)_{(i,j)} \Big)^2
\end{split}
$} 
\end{equation}

Our objective is to derive the latent feature matrix \( Q \), stored on the server, and the latent feature matrix \( P \), maintained by the sensors. To accomplish this, we utilize the Stochastic Gradient Descent (SGD)\cite{li2023nonlinear,yuan2020generalized,li2021proportional,b26,b63,b65} algorithm to address the optimization problem\cite{18,57,b43,b44}.

\subsubsection{Federated learning in the sensor}
During the \( t \)-th iteration, each sensor \( i \) downloads the latent feature matrix \( Q \) from the server along with the local latent feature vectors of other sensors within the same region. Then, sensor \( i \) in region \( h \) calculates the gradient \( \nabla (p^h_{i,.})^t \) for its local latent feature vector and the gradient \( \nabla q^t \) for \( Q \). Based on  \eqref{7}, the element-wise computations for \( \nabla (p^h_{i,.})^t \) and \( \nabla q^t_{j,.} \). They can be expressed as follows:
\begin{equation}\label{8}
\scalebox{1}{$\displaystyle 
\left\{
\begin{aligned}
\nabla {\left( {p_{i,k'}^h} \right)^t} &=- 2q_{j,k'}^t\left( {y_{i,j}} - {{\left( {p_{i,.}^h} \right)}^t}q_{j,.}^t \right) + 2\lambda {\left( {p_{i,k'}^h} \right)^t}\\
&\quad + 2{z_1}\left( {{{\left( {{L^h}{{\left( {{L^h}} \right)}^{\rm T}}} \right)}_{i,.}}{{\left( {{P^h}} \right)}^t}{{\left( {q_{j,.}^t} \right)}^{\rm T}}} \right)q_{j,k'}^t + 2 z_2\!\left( (p_{i,.}^h)^t (Q^t)^{\rm T} (DD^{\rm T})_{.,j} \right) q_{j,k'}^t, \\
\nabla q_{j,k'}^t &=- 2{\left( {p_{i,k'}^h} \right)^t}\left( {{y_{i,j}} - {{\left( {p_{i,.}^h} \right)}^t}q_{j,.}^t} \right)+ 2\lambda q_{j,k'}^t\\
&\quad + 2{z_1}\left( {{{\left( {{L^h}{{\left( {{L^h}} \right)}^{\rm T}}} \right)}_{i,.}}{{\left( {{P^h}} \right)}^t}{{\left( {q_{j,.}^t} \right)}^{\rm T}}} \right){\left( {p_{i,k'}^h} \right)^t} + 2 z_2\!\left( (p_{i,.}^h)^t (Q^t)^{\rm T} (DD^{\rm T})_{.,j} \right) (p_{i,k'}^h)^t.
\end{aligned}
\right.
$} 
\end{equation}
where \( k' \) represents the \( k' \)-th element within the latent feature dimension \( k \).

For sensor \( i \) in region \( h \), it maintains its own latent feature vector \( p^h_{i,.} \). After calculating the gradients \( \nabla (p^h_{i,.})^t \) and \( \nabla q^t \), the latent feature vector \( p^h_{i,.} \) can be updated element-wise as follows:
\begin{equation}\label{8}
{\left( {p_{i,k'}^h} \right)^{t + 1}} = {\left( {p_{i,k'}^h} \right)^t} - \eta \nabla {\left( {p_{i,k'}^h} \right)^t},
\end{equation}
where \( \eta \) denotes the learning rate of SGD, which controls the step size for each iteration of the gradient descent. After updating its local latent feature vector \( p^h_{i,.} \), sensor \( i \) transmits the gradient \( \nabla q^t \) to the server to facilitate the update of the latent feature matrix \( Q \).

\subsubsection{Federated learning in the server}
Upon receiving the gradient \( \nabla q^t \) at the \( t \)-th iteration, the server promptly updates the latent feature matrix \( Q \) using the following rule:
\begin{equation}\label{9}
    {Q^{t + 1}} = {Q^t} - \eta \nabla {q^t},
\end{equation}
So, at element-wise update of Q can be expressed as:
\begin{equation}\label{10}
    {({q_{j,k'}})^{t + 1}} = {({q_{j,k'}})^t} - \eta \nabla {({q_{j,k'}})^t},
\end{equation}
where \( (q_{j,k'})^{t+1} \) represents the \( k' \)-th element of the \( j \)-th latent feature vector in the \( t+1 \)-th iteration round.

\subsection{Privacy Analysis}
Figure~\ref{fig:FLFL} illustrates a framework where sensors can utilize their local data to train models and iteratively upload gradient information to the server in plain text format, enabling the updating of the global model. Next, we will demonstrate that a collaborative training–shared LFA model in WSNs can prevent gradient leakage according to \cite{FedMF}, thereby safeguarding sensor privacy data and protecting it from intrusion by the server. 

For sensor $i$ in region $h$, let $(p^{h}_{i,\cdot})^{t}$ denote its latent feature vector at iteration $t$. Let $U_i$ be the set of time slot observed by sensor $i$. 
The gradient with respect to the $j$-th row of $Q$ (for $j \in U_i$) that sensor $i$ sends to the server is given below.
\begin{equation}
\label{eq_15}
\begin{aligned}
G_j^t = {} & -\,2 (p_{i,.}^h)^t \big( y_{i,j} - (p_{i,.}^h)^t q_{j,.}^t \big) + 2\lambda q_{j,.}^t \\
&+ 2 z_1\!\left( \big( L^h (L^h)^{\rm T} \big)_{i,.}\, (P^h)^t\, (q_{j,.}^t)^{\rm T} \right) (p_{i,.}^h)^t \\
&+ 2 z_2\!\left( (p_{i,.}^h)^t (Q^t)^{\rm T} (DD^{\rm T})_{.,j} \right) (p_{i,.}^h)^t .
\end{aligned}
\end{equation}
Let the three auxiliary terms $\hat{y}_{i,j}^t$, $F_1^t$ and $F_2^t$ be defined as
\begin{equation}
\label{eq_16}
\begin{aligned}
&\hat{y}_{i,j}^t=(p_{i,.}^h)^t q_{j,.}^t,\\
&F_1^t = z_1\!\left( 
    \big( L^h (L^h)^{\rm T} \big)_{i,.}\,
    (P^h)^t\,
    (q_{j,.}^t)^{\rm T}
\right), \\
&F_2^t = z_2\!\left(
    (p_{i,.}^h)^t\,
    (Q^t)^{\rm T}\,
    (DD^{\rm T})_{.,j}
\right).
\end{aligned}
\end{equation}
Similarly, in the $(t+1)$-th round of iteration, we can get the following equation:
\begin{equation}
\label{eq_17}
\begin{aligned}
G_j^{t+1} = {} & -\,2 (p_{i,.}^h)^{t+1} \Big[ 
      \big( y_{i,j} - (\hat{y}_{i,j}^{t+1} \big) -\, F_1^{t+1} - F_2^{t+1} \Big]+ 2\lambda q_{j,.}^{t+1} .
\end{aligned}
\end{equation}

According to Eq.(\ref{eq_12}), the correlation between $(p^{h}_{i,\cdot})^{t}$ and $(p^{h}_{i,\cdot})^{t+1}$ is:
\begin{equation}
\label{eq_18}
\begin{aligned}
&(p_{i,.}^h)^t - (p_{i,.}^h)^{t+1}=-\,2\eta \sum_{j \in U_i} \Big\{ 
     q_{j,.}^t \Big[ \big( y_{i,j} - \hat{y}_{i,j}^t \big)-\, F_1^t - F_2^t \Big] 
     - \lambda (p_{i,.}^h)^t \Big\} .
\end{aligned}
\end{equation}

A closer look at element-wise calculation of Eq.(\ref{eq_15})-(\ref{eq_18}):
\begin{equation}
\label{eq_19}
\begin{cases}
\begin{aligned}
G_{j1}^t &= -\,2 (p_{i1}^h)^t \Big[ \big( y_{i,j} - \hat{y}_{i,j}^t \big) - F_1^t - F_2^t \Big]
           + 2\lambda q_{j1}^t, \\
&\vdots \\
G_{jk}^t &= -\,2 (p_{ik}^h)^t \Big[ \big( y_{i,j} - \hat{y}_{i,j}^t \big) - F_1^t - F_2^t \Big]
           + 2\lambda q_{jk}^t .
\end{aligned}
\end{cases}
\end{equation}

\begin{equation}
\label{eq_20}
\begin{cases}
\begin{aligned}
G_{j1}^{t+1} &= -\,2 (p_{i1}^h)^{t+1} \Big[ \big( y_{i,j} - \hat{y}_{i,j}^{t+1}\big) - F_1^{t+1} - F_2^{t+1} \Big]+ 2\lambda q_{j1}^{t+1},\\
&\vdots \\
G_{jk}^{t+1} &= -\,2 (p_{ik}^h)^{t+1} \Big[ \big( y_{i,j} - \hat{y}_{i,j}^{t+1}\big) - F_1^{t+1} - F_2^{t+1} \Big]+ 2\lambda q_{jk}^{t+1}.\\
\end{aligned}
\end{cases}
\end{equation}

\begin{equation}
\label{eq_21}
\begin{cases}
\begin{aligned}
&(p_{i1}^h)^t - (p_{i1}^h)^{t+1}=&
  \\& -\,2\eta \sum_{j \in U_i} \Big\{ q_{j1}^t \Big[ \big( y_{i,j} - \hat{y}_{i,j}^{t} \big) - F_1^t - F_2^t \Big]
       - \lambda (p_{i1}^h)^t \Big\}, \\
&\quad\quad\quad \vdots \\
&(p_{ik}^h)^t - (p_{ik}^h)^{t+1}=&
  \\& -\,2\eta \sum_{j \in U_i} \Big\{ q_{jk}^t \Big[ \big( y_{i,j} - \hat{y}_{i,j}^{t} \big) - F_1^t - F_2^t \Big]
       - \lambda (p_{ik}^h)^t \Big\}.
\end{aligned}
\end{cases}
\end{equation}

Now, let us turn to analyze the $a$-th entry of $p^{h}_{i,\cdot}$, denoted as $p^{h}_{i,a}$. According to Eq.(\ref{eq_18}), we obtain the following equations:

\begin{equation}
\label{eq_22}
\frac{(p_{ia}^h)^t}{(p_{i,(a+1)}^h)^t}
= \frac{\,G_{ja}^t - 2\lambda q_{ja}^t\,}
       {\,G_{j,(a+1)}^t - 2\lambda q_{j,(a+1)}^t\,}.
\end{equation}
\begin{equation}
\label{eq_23}
    ({y_{i,j}} - \hat{y}_{i,j}^{t}) - F_1^t - F_2^t = \frac{{G_{ja}^t - 2\lambda q_{ja}^t}}{{ - 2{{\left( {p_{ia}^h} \right)}^t}}},
\end{equation}
\begin{equation}
\label{eq_24}
{\left( {p_{ia}^h} \right)^t} - {\left( {p_{ia}^h} \right)^{t + 1}} =  - 2\eta \sum\limits_{j \in {U_i}} {\left\{ {q_{ja}^t\frac{{G_{ja}^t - 2\lambda q_{ja}^t}}{{ - 2{{\left( {p_{ia}^h} \right)}^t}}} - \lambda {{\left( {p_{ia}^h} \right)}^t}} \right\}} ,
\end{equation}
Thus, $(p^{h}_{i,\cdot})^{t+1}$ can be represented as $(p^{h}_{i,\cdot})^{t}$. 
Therefore, based on Eqs.~(\ref{eq_19}) and (\ref{eq_20}), we define, for simplicity, $O^{t}$ as
\begin{equation}
    \label{eq_25}
    O^t=(p_{ia}^h)^t
       + 2\eta \sum_{j \in U_i} \big(
           q_{ja}^t \tfrac{G_{ja}^t - 2\lambda q_{ja}^t}{-\,2 (p_{ia}^h)^t}
           - \lambda (p_{ia}^h)^t
         \big).
\end{equation}

Then we obtain the following equation:
\begin{equation}
\label{eq_26}
\begin{aligned}
&\frac{G_{ja}^t - 2\lambda q_{ja}^t}{-\,2 (p_{ia}^h)^t}
   + F_1^t + F_2^t + (p_{i,.}^h)^t q_{j,.}^t = \frac{G_{ja}^{t+1} - 2\lambda q_{ja}^{t+1}}
        {-\,2 O^t
          } + F_1^{t+1} + F_2^{t+1}+ \sum_{a=1}^k \Big(
      O^t
     \ q_{ja}^{t+1}
   \Big).
\end{aligned}
\end{equation}

Eq.~\eqref{eq_26} shows that, from the server's perspective, the observed gradients $G$ (together with $Q$) are entangled with three unknowns that remain local to the region or sensor: (i) the regional Laplacian $L_h$, (ii) the private latent vector $p^{h}_{i,\cdot}$ of sensor $i$, and (iii) the regional latent matrix $P^{h}$. 
Owing to this coupling, inverting gradients to recover the underlying sensor values is ill-posed, thereby strengthening protection against gradient-based privacy leakage.

\section{Experiments}
\subsection{General Settings}
\subsubsection{Datasets}
We evaluate on four real-world WSNs datasets spanning  CO, sea surface temperature, PM2.5, and SO$_2$. 
A summary of their characteristics is provided in Table~\ref{tab:datasets}.
\begin{table}[!h]
  \caption{The Properties of Experimental Datasets}
  \label{tab:datasets}
  \centering
  \begin{tabular}{@{}c c c c c c@{}}
    \toprule
    \textbf{No.} & \textbf{Name} & \textbf{Sensors}& \textbf{Time slots}& \textbf{Time} & \textbf{Min./Max.} \\
    \midrule
    D1 & \makecell[l]{Beijing CO\\Concentration}
       & 35 & 8647 & \makecell[l]{2014-05\\2015-04} & 0.1/20.0 \\
    D2 & \makecell[l]{Sea Surface\\Temperature}
       & 70 & 1733 & \makecell[l]{1870-01\\2014-12} & 0.01/30.31 \\
    D3 & \makecell[l]{Beijing PM2.5\\Concentration}
       & 35 & 8647 & \makecell[l]{2014-05\\2015-04} & 3.0/773.7 \\
    D4 & \makecell[l]{Chongqing SO$_2$\\Concentration}
       & 36 & 8670 & \makecell[l]{2022-01\\2022-12} & 1.0/400.0 \\

    \bottomrule
  \end{tabular}
\end{table}

\subsubsection{Evaluation Metrics}
Common evaluation metrics for assessing the accuracy of predictive models or algorithms include MAE\cite{27,b53,52,b48,b71} and RMSE\cite{26,47,b45,b46,b74}. RMSE squares residuals and therefore emphasizes large deviations, whereas MAE weights all errors equally. 
Because WSNs measurements inevitably contain outliers, reporting both metrics is informative: RMSE highlights rare large reconstruction failures, while MAE reflects typical entry-wise accuracy. MAE and RMSE are calculated as follows:
\begin{equation*}
MAE = {{\left( {\sum\limits_{{y_{i,j}} \in \gamma } {\left| {{y_{i,j}} - {{\hat y}_{i,j}}} \right|} } \right)} \mathord{\left/
 {\vphantom {{\left( {\sum\limits_{{y_{i,j}} \in \gamma } {\left| {{y_{i,j}} - {{\hat y}_{i,j}}} \right|} } \right)} {\left| \gamma  \right|}}} \right.
 \kern-\nulldelimiterspace} {\left| \gamma  \right|}},{\rm{  }}RMSE = \sqrt {{{\left( {\sum\limits_{{y_{i,j}} \in \gamma } {{{\left( {{y_{i,j}} - {{\hat y}_{i,j}}} \right)}^2}} } \right)} \mathord{\left/
 {\vphantom {{\left( {\sum\limits_{{y_{i,j}} \in \gamma } {{{\left( {{y_{i,j}} - {{\hat y}_{i,j}}} \right)}^2}} } \right)} \gamma }} \right.
 \kern-\nulldelimiterspace} \gamma }} 
\end{equation*}
where $\hat{y}_{i,j}$ denotes the estimate of $y_{i,j}$ and $\gamma$ denotes the testing set; lower values of RMSE/MAE indicate better accuracy.
\subsubsection{Baselines}
We compare the proposed FLFL model against state-of-the-art federated learning methods: RFRec\cite{RFRec},FeSoG \cite{Fsocgnn}, FedMF \cite{FedMF}, FedRec++ \cite{fedrec++}, MetaMF \cite{MetaMf}). Table \ref{tab:baselines} gives brief descriptions of these
competitors.

\begin{table}[!ht]
    \centering
    \caption{Baseline models for comparison.} 
    \label{tab:baselines} 
    \small 
    \centering
      \begin{tabular}{@{}l p{0.75\linewidth}@{}}
    \toprule
    \textbf{Model} & \textbf{Description} \\
    \midrule
    RFRec\cite{RFRec} & This federated framework efficiently optimizes the empirical risk minimization problem. (CIKM,2024) \\
    FeSoG\cite{Fsocgnn} & A federated learning framework for social recommendation using GNN, relational attention, and pseudo-labeling for privacy and personalization. (ACM TIST, 2022) \\
    FedMF\cite{FedMF} & Federated matrix factorization with homomorphic encryption. (IEEE IS 2020) \\
    FedRec++\cite{fedrec++} & Lossless federated recommendation that assigns denoising clients to remove noise in a privacy-aware manner. (AAAI 2021) \\
    MetaMF\cite{MetaMf} & Neural architecture for federated collaborative ranking via memory-based attention. (SIGIR 2020) \\
    \bottomrule
  \end{tabular}
\end{table}
\subsection{Performance Comparison}

\begin{table*}[!h]
\centering
\small
\renewcommand{\arraystretch}{1.3}
\caption{The comparison results of recovery accuracy with different sampling rates.}
\label{tab:comparison}
\begin{tabular}{>{\centering\arraybackslash}p{1cm}>{\centering\arraybackslash}p{1.3cm}>{\centering\arraybackslash}p{1.2cm}>{\centering\arraybackslash}p{1.55cm}>{\centering\arraybackslash}p{1.55cm}>{\centering\arraybackslash}p{1.55cm}>{\centering\arraybackslash}p{1.55cm}>{\centering\arraybackslash}p{1.55cm}>{\centering\arraybackslash}p{1.75cm}} 
\hline
\textbf{Dataset} & \textbf{\makecell{Sampling\\ Rate}} & \textbf{Metric}  & \textbf{RFRec}  & \textbf{FeSoG} & \textbf{FedMF} & \textbf{FedRec++} & \textbf{MetaMF} & \textbf{FLFL} \\
\hline
\multirow{6}{*}{D1} 
& \multirow{2}{*}{0.1} & MAE &$0.765_{\pm 0.026} $ & $0.434_{\pm 0.004}$ & $0.472_{\pm 0.002}$ & $0.434_{\pm 0.005}$ & $0.381_{\pm 0.001}$ & $\bm{0.305_{\pm 0.002}}$ \\
&                         & RMSE &$1.808_{\pm 0.044}$ & $0.799_{\pm 0.003}$ & $0.846_{\pm 0.002}$ & $0.782_{\pm 0.003}$ & $0.716_{\pm 0.002}$ & $\bm{0.434_{\pm 0.002}}$ \\
& \multirow{2}{*}{0.5} & MAE   & $0.309_{\pm 0.012}$ &  $0.292_{\pm 0.002}$ & $0.305_{\pm 0.002}$ & $0.287_{\pm 0.002}$ & $0.318_{\pm 0.001}$ & $\bm{0.201_{\pm 0.000}}$ \\
&                         & RMSE &$0.586_{\pm 0.017}$ & $0.578_{\pm 0.002}$ & $0.568_{\pm 0.002}$ & $0.557_{\pm 0.002}$ & $0.622_{\pm 0.001}$ & $\bm{0.399_{\pm 0.001}}$ \\
& \multirow{2}{*}{0.9} & MAE& $0.284_{\pm 0.021}$   & $0.275_{\pm 0.002}$ & $0.261_{\pm 0.001}$ & $0.250_{\pm 0.000}$ & $0.284_{\pm 0.001}$ & $\bm{0.155_{\pm 0.001}}$ \\
&                         & RMSE& $0.563_{\pm 0.033}$   & $0.568_{\pm 0.002}$ & $0.529_{\pm 0.002}$ & $0.512_{\pm 0.000}$ & $0.565_{\pm 0.001}$ & $\bm{0.328_{\pm 0.001}}$ \\
\hline
\multirow{6}{*}{D2} 
& \multirow{2}{*}{0.1} & MAE   & $0.535_{\pm 0.024}$ &$1.477_{\pm 0.010}$ & $0.922_{\pm 0.018}$ & $0.956_{\pm 0.016}$ & $0.438_{\pm 0.008}$ & $\bm{0.324_{\pm 0.002}}$ \\
&                         & RMSE& $0.889_{\pm 0.026}$  & $2.132_{\pm 0.016}$ & $1.588_{\pm 0.020}$ & $1.197_{\pm 0.019}$ & $0.591_{\pm 0.010}$ & $\bm{0.433_{\pm 0.011}}$ \\
& \multirow{2}{*}{0.5} & MAE  & $0.172_{\pm 0.019}$  & $0.622_{\pm 0.006}$ & $0.187_{\pm 0.002}$ & $0.198_{\pm 0.001}$ & $0.264_{\pm 0.004}$ & $\bm{0.157_{\pm 0.001}}$ \\
&                         & RMSE  & $0.239_{\pm 0.022}$ & $0.852_{\pm 0.007}$ & $0.253_{\pm 0.001}$ & $0.260_{\pm 0.002}$ & $0.341_{\pm 0.006}$ & $\bm{0.215_{\pm 0.001}}$ \\
& \multirow{2}{*}{0.9} & MAE  & $0.133_{\pm 0.014}$ & $0.565_{\pm 0.003}$ & $0.129_{\pm 0.001}$ & $0.126_{\pm 0.001}$ & $0.192_{\pm 0.004}$ & $\bm{0.120_{\pm 0.001}}$ \\
&                         & RMSE  & $0.179_{\pm 0.018}$ & $0.738_{\pm 0.003}$ & $0.173_{\pm 0.001}$ & $0.169_{\pm 0.001}$ & $0.250_{\pm 0.004}$ & $\bm{0.161_{\pm 0.001}}$ \\
\hline
\multirow{6}{*}{D3} 
& \multirow{2}{*}{0.1} & MAE  & $36.272_{\pm 0.621}$  & $26.256_{\pm 0.653}$ & $27.024_{\pm 0.043}$ & $23.697_{\pm 0.053}$ & $25.607_{\pm 0.053}$ & $\bm{18.023_{\pm 0.290}}$ \\
&                         & RMSE& $70.306_{\pm 0.642}$ &  $46.603_{\pm 0.692}$ & $48.078_{\pm 0.033}$ & $41.243_{\pm 0.028}$ & $43.126_{\pm 0.064}$ & $\bm{30.479_{\pm 0.128}}$ \\
& \multirow{2}{*}{0.5} & MAE   & $17.663_{\pm 1.057}$ & $19.461_{\pm 0.752}$ & $17.791_{\pm 0.033}$ & $16.859_{\pm 0.018}$ & $18.631_{\pm 0.069}$ & $\bm{10.280_{\pm 0.006}}$ \\
&                         & RMSE  & $29.977_{\pm 1.076}$ & $34.427_{\pm 0.649}$ & $30.807_{\pm 0.032}$ & $29.896_{\pm 0.016}$ & $32.557_{\pm 0.099}$ & $\bm{19.335_{\pm 0.023}}$ \\
& \multirow{2}{*}{0.9} & MAE   & $15.630_{\pm 1.468}$  & $18.722_{\pm 0.559}$ & $14.486_{\pm 0.012}$ & $14.831_{\pm 0.011}$ & $16.548_{\pm 0.041}$ & $\bm{7.993_{\pm 0.001}}$ \\
&                         & RMSE  & $26.425_{\pm 0.938}$ & $32.675_{\pm 0.571}$ & $25.861_{\pm 0.015}$ & $26.014_{\pm 0.012}$ & $28.389_{\pm 0.114}$ & $\bm{14.741_{\pm 0.001}}$ \\
\hline

\multirow{6}{*}{D4} 
& \multirow{2}{*}{0.1} & MAE    & ${7.320_{\pm 0.012}}$ & $5.268_{\pm 0.830}$ & $3.921_{\pm 0.023}$ & $3.753_{\pm 0.027}$ & $3.274_{\pm 0.014}$ & $\bm{2.307_{\pm 0.029}}$ \\
&                         & RMSE  & $18.102_{\pm 0.017}$ &  $7.956_{\pm 0.712}$ & $6.713_{\pm 0.019}$ & $6.526_{\pm 0.022}$ & $5.919_{\pm 0.004}$ & $\bm{4.806_{\pm 0.024}}$ \\
& \multirow{2}{*}{0.5} & MAE & $3.511_{\pm 0.005}$  & $5.130_{\pm 0.058}$ & $3.548_{\pm 0.009}$ & $3.330_{\pm 0.007}$ & $3.070_{\pm 0.007}$ & $\bm{1.830_{\pm 0.002}}$ \\
&                         & RMSE& $8.511_{\pm 0.007}$ &  $7.955_{\pm 0.054}$ & $6.146_{\pm 0.010}$ & $5.821_{\pm 0.008}$ & $5.701_{\pm 0.006}$ & $\bm{4.114_{\pm 0.003}}$ \\
& \multirow{2}{*}{0.9} & MAE & $3.692_{\pm 0.008}$ &  $4.837_{\pm 0.042}$ & $3.523_{\pm 0.009}$ & $3.382_{\pm 0.006}$ & $3.122_{\pm 0.007}$ & $\bm{1.780_{\pm 0.006}}$ \\
&                         & RMSE & $6.023_{\pm 0.016}$ &  $7.787_{\pm 0.046}$ & $6.123_{\pm 0.008}$ & $5.973_{\pm 0.005}$ & $5.756_{\pm 0.008}$ & $\bm{4.000_{\pm 0.008}}$ \\
\hline


\multicolumn{2}{c}{\multirow{3}{*}{\shortstack{\textbf{Statistical} \\ \textbf{Analysis}}}} 
 & \textbf{Win/Loss} & \textbf{24/0} & \textbf{24/0} & \textbf{24/0} & \textbf{24/0}& \textbf{24/0} & \textbf{120/0} \\
\multicolumn{2}{c}{~} 
 & \textbf{p-value} & \textbf{0.0001} & \textbf{0.0001}  & \textbf{0.0001}& \textbf{0.0001} & \textbf{0.0001} & \textbf{--} \\
\multicolumn{2}{c}{~} 
 & \textbf{F-rank}  & \textbf{4.376} & \textbf{5.229} & \textbf{3.883} & \textbf{2.813} & \textbf{3.688} & \textbf{1} \\
\hline
\end{tabular}
\end{table*}
\begin{figure*}[t]
    \centering
    \includegraphics[width=0.9\linewidth]{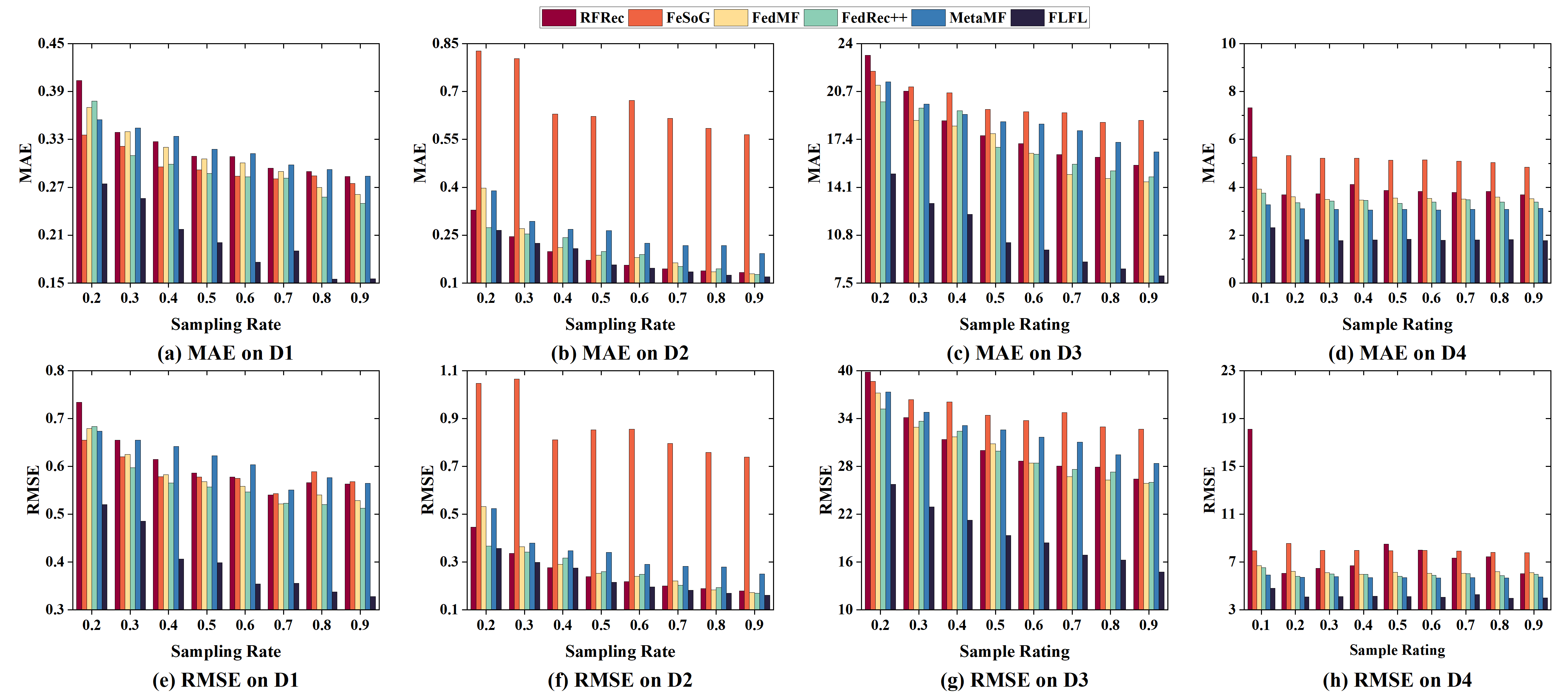}
    \caption{The comparison results of recovery accuracy with different sampling rates from 0.1 to 0.9 on all datasets.}
    \label{fig:comparison}
\end{figure*}
This set of experiments gradually increases the sampling rate from 0.1 to 0.9 to compare the performance of all involved models. The comparison results across all datasets are shown in Table ~\ref{tab:comparison} and Fig \ref{fig:comparison}. To further interpret these results, statistical analyses including win/loss counts, the Wilcoxon signed-ranks test \cite{45,b51}, and the Friedman test \cite{38} are performed. The win/loss analysis counts the number of cases where FLFL achieves lower or higher RMSE compared to each baseline model as the sampling rate changes. The Wilcoxon signed-ranks test, a nonparametric pairwise comparison method, assesses whether FLFL-SSR significantly outperforms each comparison model by examining the p-value's significance level. Meanwhile, the Friedman test compares the performance of multiple models across various scenarios simultaneously, using the F-rank value. A lower F-rank indicating superior recovery accuracy.

As shown in Table~\ref{tab:comparison}, both MAE and RMSE generally decrease as the sampling rate increases. 
Two additional findings are noteworthy:

\textbf{Relative performance.} Compared to federated baselines, FLFL achieves the lowest MAE / RMSE in all settings evaluated.

\textbf{Statistical significance.} Across the four datasets, all $p$-values are below $0.05$, and FLFL attains the best (lowest) Friedman rank, indicating statistically significant improvements over federated  models.

Overall, FLFL sustains higher accuracy than non-federated spatio-temporal methods and delivers the best recovery performance among federated approaches, attributable to its explicit modeling of spatio-temporal correlations in WSNs.

\subsection{Ablation Study}
\begin{figure}[!h]
    \centering
    \includegraphics[width=0.9\linewidth]{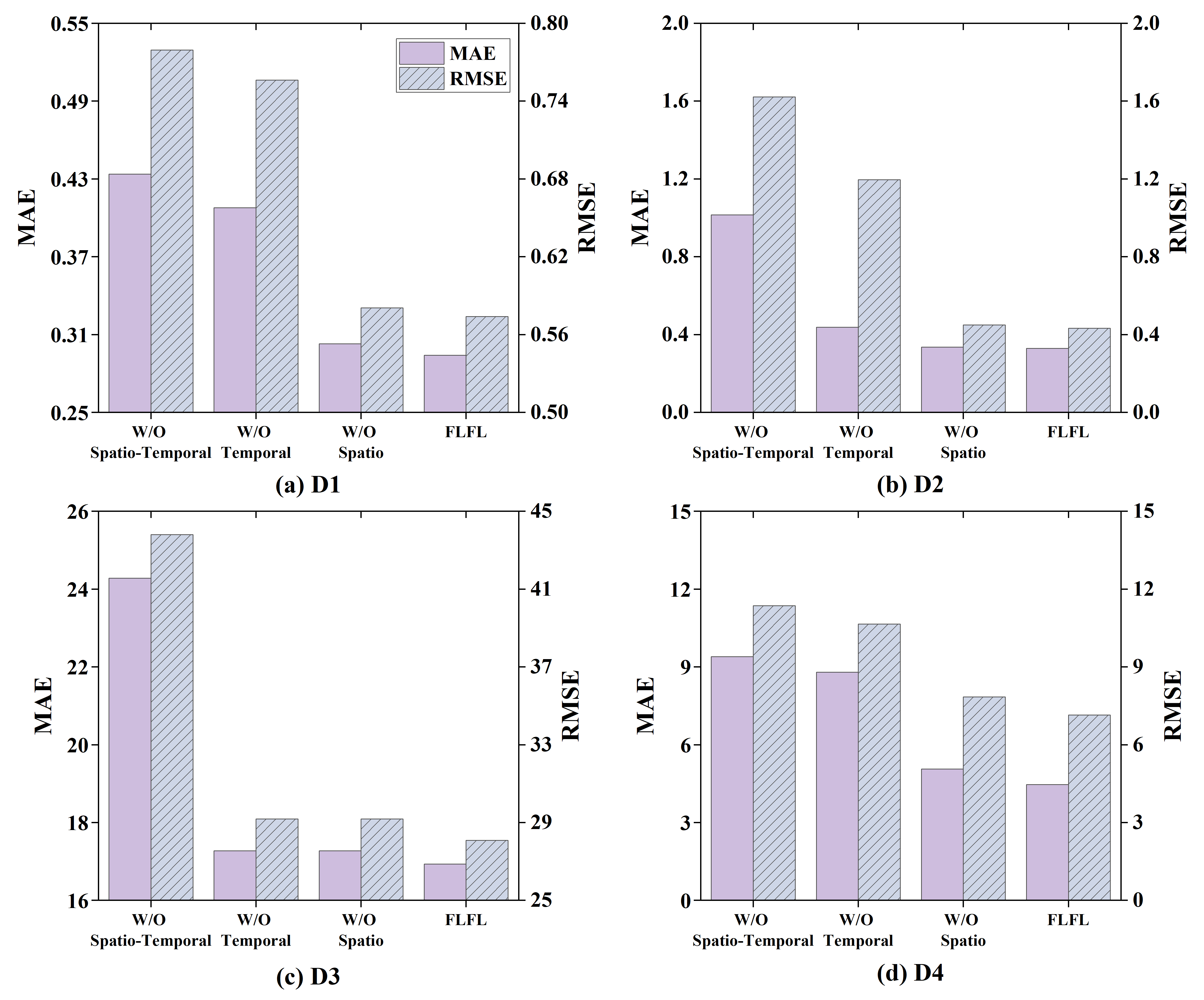}
    \caption{The Ablation study of Spatio-Temporal Correlations}
    \label{fig:st}
\end{figure}
\noindent Under a fixed sampling rate of $0.1$, we perform a systematic ablation of spatio-temporal correlations in FLFL to assess their contribution to recovery accuracy. Fig~\ref{fig:st} reports results for four settings: no spatio-temporal correlation, spatial-only, temporal-only, and joint spatio-temporal modeling. The experiments show that introducing either spatial or temporal correlation alone markedly improves recovery accuracy over the baseline without any correlations; and jointly modeling spatio-temporal correlations achieves the best performance across all datasets, significantly surpassing the single-correlation variants. In sum, explicitly modeling spatio-temporal correlations in FLFL is both necessary and effective for improving missing-data recovery.

\section{CONCLUSION}
This paper proposes FLFL, for privacy-preserving spatio\mbox{-}temporal signal recovery in WSNs. The core idea is twofold: (i) a sensor-level federated learning framework based on LFL, in which each sensor uploads gradient information rather than raw data to train a privacy-preserving recovery model; and (ii) incorporation of spatio\mbox{-}temporal correlation into the federated framework as a regularization constraint to enhance recovery accuracy. This design endows FLFL with both strong privacy protection and high accuracy for WSNs data recovery. Experiments on four real-world WSNs datasets demonstrate that FLFL significantly outperforms related state-of-the-art models in terms of both recovery accuracy and privacy preservation. In future work, we will further optimize FLFL to address outliers arising in WSNs data collection and to improve the model’s robustness.

\bibliographystyle{unsrt}  
\bibliography{references}

\end{document}